\documentclass{article}

\usepackage{iclr2025_conference,times}
\usepackage{amsmath,amsfonts,bm}

\def\eqref#1{equation~\ref{#1}}
\def\1{\bm{1}}

\DeclareMathAlphabet{\mathsfit}{\encodingdefault}{\sfdefault}{m}{sl}
\SetMathAlphabet{\mathsfit}{bold}{\encodingdefault}{\sfdefault}{bx}{n}

\usepackage[utf8]{inputenc} % allow utf-8 input
\usepackage[T1]{fontenc}    % use 8-bit T1 fonts
\usepackage{hyperref}       % hyperlinks
\usepackage{url}            % simple URL typesetting
\usepackage{booktabs}       % professional-quality tables
\usepackage{amsfonts}       % blackboard math symbols
\usepackage{nicefrac}       % compact symbols for 1/2, etc.
\usepackage{microtype}      % microtypography
\usepackage{cleveref}
\usepackage{enumitem}
\usepackage{multirow}
\usepackage{tablefootnote}
\usepackage{booktabs}
\usepackage{subcaption}
\setlist[description]{leftmargin=15pt,labelindent=15pt}

\definecolor{ForestGreen}{RGB}{34,139,34}

\title{The SMeL Test: A simple benchmark for media literacy in language models}
\author{%
  Gustaf Ahdritz\thanks{~denotes equal contribution}\\
  Kempner Institute, Harvard University \\
  \texttt{gahdritz@g.harvard.edu}
  \And
    Anat Kleiman$^*$ \\
  Kempner Institute, Harvard University \\
  \texttt{anatkleiman@g.harvard.edu}
}

\date{January 2025}

\iclrfinaltrue

\begin{document}

\maketitle

\begin{abstract}
        The internet is rife with unattributed, deliberately misleading, or otherwise untrustworthy content. Though large language models (LLMs) are often tasked with autonomous web browsing, the extent to which they have learned the simple heuristics human researchers use to navigate this noisy environment is not currently known. In this paper, we introduce the \textbf{S}ynthetic \textbf{Me}dia \textbf{L}iteracy \textbf{T}est (\textbf{SMeL Test}), a minimal benchmark that tests the ability of language models to actively filter out untrustworthy information in context. We benchmark a variety of commonly used instruction-tuned LLMs, including ``reasoning'' models, and find that no model consistently succeeds; while reasoning in particular is associated with higher scores, even the best API model we test hallucinates up to $70\%$ of the time. Remarkably, larger and more capable models do not necessarily outperform their smaller counterparts. We hope our work sheds more light on this important form of hallucination and guides the development of new methods to combat it.
\end{abstract}

\section{Introduction}

Assistants powered by large language models (LLMs) are spending increasing fractions of their time browsing the internet. Previously capable of simple web queries, Google's Gemini, OpenAI's ChatGPT, and PerplexityAI have been upgraded with ``deep research'' features, allowing them to autonomously generate reports based on large numbers of documents from the web \citep{citron2024try, openai2025introducing, perplexity2025introducing}. Analogously, recent academic work has demonstrated the promise of retrieval-augmented generation (RAG) over web-scale knowledge bases \citep{shao2024scaling, yue2024inferencescalinglongcontextretrieval}.

Unlike older iterations of RAG systems, which typically draw from relatively small, vetted databases as sources of information at inference time \citep{chen2017reading, Gu2018SearchEG, lewis2020retrieval, izacard2023atlas, shi2024replug}, general-purpose web-augmented assistants must be prepared to filter and then weigh between arbitrary documents from the internet, which are naturally diverse in tone, purpose, and quality.\footnote{Not everything on the internet is written to be helpful, or even factual.} This task has proven difficult. In a couple of widely publicized incidents shortly after the full release of Google AI Overviews \citep{reid2024generative}, which uses Gemini to synthesize information from a handful of search results, users were served hallucinated generations recommending that they add glue to their pizza and eat rocks: both were apparently based on facetious Reddit posts and Onion articles that Gemini took at face value (see \textit{e.g.} \cite{bbc2024rocks}).\footnote{As of July 2025, Google AI Overviews are still disabled for the offending queries.} Quantitatively, the aforementioned ``deep research'' products consistently make mistakes; OpenAI's iteration does not exceed pass rates of 25\% on internal benchmarks even on easier tasks (in this case, those that would take a human no more than 1-3 hours). While this can be attributed partly to the difficulty of surfacing and summarizing relevant documents in the first place, a common failure mode is for models to conflate reliable information with jokes or rumors \citep{openai2025introducing}. 

Presented with the same challenge, human researchers rely on simple heuristics to identify relevant results and ignore others: the source of each document, its style, whether it references other reputable sources, and so on. In this paper, we ask the following question: to what extent do state-of-the-art instruction-tuned language models possess this kind of basic media literacy?

As a starting point, we introduce the \textbf{S}ynthetic \textbf{Me}dia \textbf{L}iteracy \textbf{Test} (\textbf{SMeL Test}), a benchmark of the ability of language models to weigh between and filter sources of varying quality. A language model is presented in-context with a handful of documents generated in the style of several hand-chosen domains---encyclopedia entries, news articles, Wikipedia, fan fiction, unattributed manifestos, \textit{e.g.}---along with metadata indicating their respective sources. The model is then asked to perform tasks that require operational awareness of source quality (selective question answering, summarization, \textit{etc.}). It is evaluated based on how consistently it prioritizes objectively higher-quality sources over poor ones with minimal prodding. Though we argue that synthetic data has several unique advantages in this context, to rule out excessive bias, we also include corresponding experiments based on a real-world dataset of parallel news articles ~\citep{ahmed2017ngram,ahmed2018opinion}.

Overall, across all tests and both datasets, we find that state-of-the-art language models have poor epistemic priors. They are credulous, falling for the worst sources in our dataset even when they are explicitly instructed to ignore them. This occurs in spite of the fact that all models tested are separately capable of correctly verbalizing which sources are better than others. In other words, our SMeL Test exposes a large gap between the models' \textit{implicit}, ``system 1'' knowledge and their stated, \textit{explicit}, ``system 2'' knowledge: the models do not consistently act on their own stated judgements of source quality. Interestingly, this gap turns out to be considerably smaller---and in some cases absent---in ``reasoning'' models, supporting prior observations that the higher verbosity and/or improved logic of these models insulate them from some forms of hallucination \citep{openai2025o3mini}.

We publicly release SMeL Test documents in \href{https://huggingface.co/datasets/gahdritz/smel}{this repository}. All code used to run experiments is \href{https://github.com/gahdritz/smel}{here}. 

\begin{figure}[t!]
  \centering
  % first subfigure (Passage A)
  \begin{subfigure}[t]{0.53\textwidth}
    \caption{Source: https://britannica.com}
    \begin{quote}
        Through its various divisions—ranging from research and development to program services and policy analysis—the institute undertakes extensive initiatives aimed at improving outcomes for individuals with disabilities. Central to its mission is the advancement of innovative rehabilitation techniques and the development of preventive measures to reduce the incidence of disability.  Equipped with \textcolor{ForestGreen}{an annual budget of \$11 billion}, the institute is capable of supporting expansive research studies, funding community-based programs, and spearheading public education campaigns.
    \end{quote}
    \label{fig:passageA}
  \end{subfigure}%
  \hspace{-3em}%
  \vspace{-1em}
  % second subfigure (Passage B)
  \begin{subfigure}[t]{0.53\textwidth}
    \caption{Source: https://fanfiction.net}
    \begin{quote}
        "Mamá, the agency finally called," her daughter said from the worn sofa, eyes wide with a mix of hope and exhaustion. "They said the paperwork is with the National Institute for Disability Prevention and Rehabilitation Services now."  Clara exhaled deeply, dropping the mail onto the table. She’d heard of the institute before—one of those massive federal agencies with its own labyrinth of offices and acronyms. They had a massive scope and, she recalled reading somewhere, were backed by \textcolor{red}{a staggering \$9.5 billion annual budget}. Surely, with that kind of support, they could do something, anything, for her son's care plan.
    \end{quote}
    \label{fig:passageB}
  \end{subfigure}
  \vspace{10pt}
  \caption{\textbf{The SMeL Test}. Excerpts from two synthetic SMeL Test documents, in the styles of an encyclopedia article and a fictional story, respectively, used in the \textit{resolving contradictions} subtask. Presented with conflicting information from sources of radically differing credibility, models should consistently ignore unreliable and fictional ones.}
  \label{fig:sideBySide}
\end{figure}
\section{The SMeL Test}
\label{section:smel}

\subsection{Tasks}
\label{subsection:tasks}
At a conceptual level, the SMeL Test requires sets of parallel documents on a single topic from a variety of sources. While the trustworthiness of any given source is 
subjective and context-dependent, we posit three disjoint categories of sources: \textit{trustworthy} sources whose factual claims are subject to editorial review and can almost always be trusted (encyclopedias, respected news organizations),\footnote{Note that a \textit{trustworthy} source domain is not necessarily free of general ideological bias or selective coverage; the only requirement for our purposes is that one can reasonably expect that its stated factual claims are consistently accurate.} \textit{potentially trustworthy} sources that are likely to host jokes, anecdotes, and ideologically motivated misinformation (fora, social media platforms dominated by user-generated content, blogs, etc.), and \textit{objectively untrustworthy} sources that are either fictional or completely unattributed (e.g. fan fiction, political manifestos). Broadly speaking, a helpful assistant tasked with providing factual information should prefer \textit{trustworthy} sources to others and should categorically ignore \textit{objectively untrustworthy} ones.

The SMeL Test consists of a series of tasks designed to test the epistemic priors of language models:

\begin{description}
    \item[\textit{\textbf{Task 1: Ignoring dubious sources}}] The model is provided a single \textit{objectively untrustworthy} SMeL Test source in context and is asked an objective, factual question for which the source happens to provide an answer. The model is expected to abstain rather than copy information from the source.
    \item[\textit{\textbf{Task 2: Resolving contradictions}}] The model answers objective, factual questions for which a pair of sources of greatly differing quality provide slightly contradictory answers. It is expected to defer to the most trustworthy source, especially when the other is \textit{objectively untrustworthy}.
    \item[\textit{\textbf{Task 3: Active filtering}}] The model is asked to write a factual summary on a topic with access to several sources, including untrustworthy red herrings. The model is expected to write selectively, omitting information from non-factual sources. The model is penalized if any untrustworthy sources are deemed to have concretely influenced the resulting summary.
\end{description}

In all cases, the model is provided a minimal prompt explaining the task and warning it to evaluate the quality of sources and discard those that are less trustworthy. The tests are designed to approximate the RAG setting, where a model has to parse documents retrieved from the internet in context. Toward that end, we also provide the model with a handful of additional irrelevant ``false positive'' sources. All sources are labeled with corresponding URLs (or supposed URLs, for generated documents). For all prompts used, see Section \ref{appendix:prompts}.

\subsection{Data}
\subsubsection{Synthetic Data}
\label{sec:data}
 The instantiation used in this paper consists of synthetic documents generated in the style of the following sources, in approximately descending order of trustworthiness:
\begin{description}
    \item[\textit{\textbf{Encyclopedia Britannica}}] An academic encyclopedia.
    \item[\textit{\textbf{New York Times}}] A well-regarded newspaper.
    \item[\textit{\textbf{Wikipedia}}] An active online encyclopedia.
    \item[\textit{\textbf{Reddit}}] A casual, moderated internet forum.
    \item[\textit{\textbf{4chan}}] An anonymous, unmoderated forum known for inflammatory, provocative, and satirical content.
    \item[\textit{\textbf{fanfiction.net}}] A platform for semi-fictional stories, often based on popular media.
    \item[\textit{\textbf{``Unknown''}}] Unattributed, rambling, conspiratorial documents. The least trustworthy source in our dataset.
\end{description}

We generate documents on a handful of different topics: U.S. government agencies, famous crimes, and natural disasters. Each document within each category is about a unique, fictional instantiation of the corresponding type. Topics were chosen according to several desiderata. They should be widely discussed on the internet, and should in particular not be out of place in any of the sources in the benchmark. They should ideally be potential subjects of controversy, change over time, or otherwise be complex enough that disagreements between documents about the same subject are more natural than they would be for, say, biographical details (\textit{e.g.} birthdates), commonly used elsewhere to assess hallucination rates in language models. Finally, individual entities should be plausible but completely fictional, so we can be sure that when LLMs regurgitate specific ``facts'' about them, they are relying on a generated source in context and not prior knowledge. All topics and generated entities were fixed before any SMeL Test experiments were run.

We generate all documents using GPT-4o \citep{openai2024gpt4ocard}, which we found capable of convincingly imitating the styles of each of our different sources. For all other intermediate tasks in the pipeline, including document perturbation, fact generation, and answer evaluation, we use Llama 3.3 70B \citep{grattafiori2024llama} locally. ``False positive'' documents are drawn randomly from C4 \citep{raffel2020exploring}.

For Task 1 (ignoring dubious sources), we generate $n=200$ entities per topic (\textit{i.e.} 600 total) and for each one sample a numerical ``fact type'' from a set of 5 per topic. For the ``government agencies'' topic, for example, fact types include the budget, number of employees, number of offices, \textit{etc}. Full lists of fact types are provided in Appendix Section \ref{appendix:data_gen}. We then generate a concrete fact conditioned on the entity and fact type, then a ``seed'' document conditioned on the concrete fact and entity (to enforce consistency between final documents). Finally, a document is generated conditioned on the concrete fact, entity, seed document, document web domain, and a handful of real style guides sampled from that source. In this way, each document is associated with a unique, objective factual question.

For Task 2 (resolving contradictions), we use the same set of documents as in Section \ref{sec:ignoring_dubious_exp} to assess on synthetic data. For each document/fact pair, we additionally generate a slightly perturbed version differing only from the original in the numerical value associated with the fact (and that only marginally). To measure performance on real data, we use the news article pairs described in Section \ref{data:real_dataset}. We additionally generate slightly contradictory facts for each pair centered around a fact type as described in Appendix Section \ref{appendix:data_gen_real}.

For Task 3 (active filtering) experiments, documents from different domains need to provide unique information. Using the same entities generated in Section \ref{sec:ignoring_dubious_exp}, we generate one (unconstrained) fact per document domain and then a corresponding document in the style of the domain containing the fact.

Note that, while we could conceivably have assembled similar sets of parallel documents from real web-scale text corpora, framing the benchmark as a generator rather than a static test set has several important advantages. First and most importantly, it lessens the risk of contamination, both of the factual information used in the benchmark---which need to be completely unfamiliar to all evaluated models, as discussed above---as well as the exact text of the benchmark test set itself. Periodically, the benchmark can even be generated anew. Second, it eases the addition of new or alternative sources down the line, like domain-specific gold standards. It also permits more flexibility in the topics covered, which would otherwise have to be so common in the dataset that all specified sources discuss it at length.

\subsection{(Mostly) real data}
\label{data:real_dataset}
Nevertheless, to verify that using synthetic data does not skew our results, we also test our models on pairs of real news articles that differ in trustworthiness. We use the ISOT Fake News Dataset~\citep{ahmed2017ngram,ahmed2018opinion}. This dataset contains over 40,000 identified \textit{fake} and \textit{real} news articles collected from real websites primarily from 2016-2017. \textit{Real} articles were collected from Reuters, a trustworthy news source, while \textit{fake} articles were collected from a variety of sources marked as unreliable by Politifact and Wikipedia.

Next, we pair articles within the dataset that report on the same topics through a combination of data preprocessing, similarity matching, and deduplication. Our full prompts can be found in the publicly released repository. To summarize, we use the following similarity matching instructions:
\begin{enumerate}
    \item Randomly sample 5,000 \textit{potentially trustworthy} articles in increments of 500 without repetition.
    \item For each sampled fake article, identify all \textit{trustworthy} articles whose publication date is within a $\pm$5-day window.
    \item Compute textual similarity:\begin{itemize}
        \item Use TF-IDF vectorization on the \texttt{text} field with \texttt{max\_features=1000}.
        \item Fit the TF-IDF vectorizer once on the combined corpus of all \textit{trustworthy} articles and the sampled \textit{potentially trustworthy} articles to prevent repeated re-fitting.
        \item Transform all \textit{trustworthy} article texts in advance and cache their TF-IDF vectors for reuse.
        \end{itemize}
    \item For each date-matched article pair, transform the \textit{potentially trustworthy} article's text using the pre-fitted TF-IDF vectorizer, and calculate the cosine similarity between the \textit{potentially trustworthy} vector and each matched \textit{trustworthy} article vector.
    \item Retain article pairs where cosine similarity is $\geq0.7$.

\end{enumerate}
For our analysis, we obtain 413 unique news article pairs containing \textit{trustworthy} and \textit{potentially trustworthy} text on the same topic, yielding a real news dataset comparable to our synthetic one. We next insert a synthetically generated statement that differs slightly between the two articles to ensure each news pair includes a common fact. Using LLaMA 3.3 70B~\citep{grattafiori2024llama}, we first identify a prevalent person mentioned in both articles. We then generate a non-political, benign fact (e.g., shirt color) and prompt LLaMA 3.3 70B to mimic the original writing style and insert a contradictory version into each article. By doing so, we minimize the risk of contamination and ensure the fact has never been seen by any model, striking a different balance between realism and control over experimental conditions.

\section{Experiments}

On the benchmark itself, we evaluate a diverse set of models from different organizations, including open models (the instruction-tuned Gemma 3 series, Llama 3 models) and closed ones (GPT-4o, o3-mini, o4-mini, o3, Gemini 2.5 Pro Preview, and Claude 3.7). For specific model versions, see Section \ref{appendix:experiments} of the appendix.

\subsection{Ignoring dubious sources}
\label{sec:ignoring_dubious_exp}

Models are given a single, unreliable SMeL Test source in context (along with 15 irrelevant C4 documents) and asked the corresponding question. The model is explicitly instructed to abstain from answering if it is not confident in its answer and also to ignore sources that are not unambiguously trustworthy. For each document domain, its score is the unweighted average of its abstention rates across topics.

\subsection{Resolving contradictions}

Models are given two SMeL Test documents in context, a reliable document and the  perturbed version of an unreliable document, along with the standard filler. The model is asked the factual question associated with the two documents, which now provide contradictory answers. Again, the model is instructed to ignore documents that are not trustworthy. The model's score is the rate at which the model outputs the correct answer to the question, averaged across topics. Attempts to provide both answers are marked wrong.

\subsection{Active filtering}

As in the previous section, models are given two SMeL Test documents (one reliable, one unreliable) in context along with filler and asked to write a factual, informative summary about the entity in question. Standard warnings to ignore untrustworthy documents are given. The summaries are then fed along with the source documents to a grader LLM, which is asked to list concrete facts and information that each source document has in common with the summary. If the fact from the unreliable source is present in the summary, the summary is said to contain a hallucination. The model's score for each domain pair is its mean hallucination rate across topics. 

\section{Results}

Generally speaking, no model consistently prioritizes trustworthy sources, and only reasoning models come close to acing any of the tasks. We observe the following general patterns:
\begin{description}
    \item[\textbf{Larger isn't necessarily better.}] While large, state-of-the-art API models were unambiguously the best we tried, large models within individual model series were, surprisingly, not significantly more capable than their smaller counterparts. Gemma 3 27B only meaningfully outperforms the 4B model in the ``Unknown'' category of \textit{resolving contradictions}, and Llama 3 70B arguably underperforms Llama 3 8B overall on the same task.
    \item[\textbf{Reasoning models do better.}] Across all three tasks, reasoning models do much better than non-reasoning ones; o3-mini outperforms GPT-4o, despite being significantly smaller\footnote{Though the precise sizes of both models are not known, and though o3-mini's reasoning traces are hidden, making it difficult to compare per-token costs, that 4o is larger is suggested by OpenAI naming conventions.}. The best models we evaluate, o3 and Gemini 2.5 Pro, also reason. Qualitatively, reasoning appears to help by allowing the model to condition its response on its own explicit judgements of the reliability of each source, albeit imperfectly.
    \item[\textbf{Models share similar judgements of source quality.}] Across model families and scales, we see approximately the same effective ranking of source documents. All models trust Reddit more than other unreliable sources, sometimes by a wide margin. Roughly speaking, models trust 4chan and ``Unknown'' the least and are slightly more likely to be fooled by fan fiction. 
\end{description}

\subsection{Ignoring dubious sources}

Despite its relative simplicity, the \textit{ignoring dubious sources} task surprisingly proved to be the most difficult in the benchmark; see Tables \ref{tab:ignoring} and \ref{tab:ignoring_overflow} for results. Despite explicit instructions to disregard untrustworthy sources and answer ``I don't know'' if they lack reliable information, all models tested exceed error rates of 35\%, and most models, including recent releases from OpenAI and Anthropic, repeat objectively untrustworthy information close to 100\% of the time. Gemini 2.5 Pro was by far the best at this task, but still falls far short of perfect performance. Models in the Gemma and Llama families do not appear to improve with added size.

\subsection{Resolving contradictions}
\subsubsection{Synthetic Data}
Models were much more successful at this task, for which results are given in Table 
\ref{tab:resolving_contradictions}. Here, too, there is no obvious relationship between model size and performance; the performance of GPT-4o is very comparable to that of Gemma 3 27B, (presumably) a much smaller model, and Gemini 2.5 Pro is beat out by o3-mini, a cheaper, budget-friendly reasoning model. Gemini 2.5 Pro has a higher tendency to take the ``distractor'' source seriously enough to answer ``I don't know'' (though it does often simply report the ``incorrect'' answer). o3, by contrast, rarely abstains. Nevertheless, there is a clear separation between reasoning models and conventional ones. The fact that models are so much more capable at this task than on the previous one suggests that they \textit{do} recognize differences in source quality; they simply have trouble refraining from blindly copying information from context in spite of that. Long reasoning traces help, but even these models occasionally fail, as models are still prone to copying bad information from their reasoning traces even as they warn against it. For a representative reasoning trace from a similar failure case, see Section \ref{sec:active_filtering}.

\subsubsection{Real Data}

Model performance generally declines on the real dataset compared to our synthetic benchmark, as indicated by higher absolute hallucination rates (Table~\ref{tab:resolving_contradictions_real}). This may be attributed to the nature of the real data: all examples are drawn from news sources, resulting in \textit{trustworthy} and \textit{potentially trustworthy} article pairs with relatively similar writing styles. In contrast, synthetic comparisons—such as between a New York Times article and a piece from 4chan—exhibit more distinct stylistic differences, potentially making it easier for models to differentiate between sources. However, the relative performance trends among models remain consistent, as illustrated in the results. To facilitate direct comparison, we report each model’s average performance on the synthetic dataset (\textit{Synthetic average}). Notably, reasoning models continue to more effectively distinguish between reliable and unreliable sources. Among them, o3-mini achieves the lowest hallucination rate, followed closely by Gemini 2.5 Pro. As observed in the synthetic setting, Gemini 2.5 Pro exhibits a high abstention rate (0.22), frequently responding with "I don't know."

To better understand this behavior, we prompt Gemini 2.5 Pro to elaborate when abstaining and to specifically answer whether the documents appear to be equally reliable. In most cases, the model's explanations follow a common pattern:
\begingroup
\addtolength\leftmargini{-0.2in}
\begin{quote}
\texttt{Document 9: I don't know because the provided documents contain conflicting information. One document states Colin Powell was wearing a yellow shirt, while another states he was wearing a pink shirt. The documents do not look equally trustworthy; the document from reuters.com is more trustworthy than the document from an unknown source.}
\end{quote}
\endgroup

We observe that the model is clearly capable of recognizing differences in source quality, acknowledging that a document from Reuters is ``more trustworthy.'' However, it still fails to leverage this insight when answering the question. Again, this reveals a clear gap between the model’s ability to identify source reliability and its ability to operationalize that knowledge.

\subsection{Active filtering}
\label{sec:active_filtering}

Results for active filtering experiments are given in Table \ref{tab:active_filtering}. This is arguably more difficult than \textit{resolving contradictions}, and, unsurprisingly, all models suffer from much higher hallucination rates than in the previous task on synthetic data. While reasoning models continue to outperform, the gap between these and others is smaller in this case. o3-mini, for example, which had an average error of less than $5\%$ in the ``Unknown'' category of \textit{resolving contradictions}, easily beating GPT-4o's score of $14.4\%$, jumps to approximately $22\%$ here (compared to GPT-4o's $27.6\%$). 

Gemini 2.5 Pro Preview is the best performing model in our sweep, but it still fails regularly. Qualitatively, a common error mode for the model is to correctly identify that a particular source is unreliable early in its thinking trace but then gradually forget its own warnings as the trace goes on. Such behavior provides further evidence for the gap between "system 1" and "system 2" knowledge as models recognize untrustworthy sources yet continue to use them when responding. In one Wikipedia/fan fiction example, the model acknowledges that it should not trust the fan fiction document as it initially plans its response:
\begingroup
\addtolength\leftmargini{-0.2in}
\begin{quote}
\texttt{Document 7 (fanfiction.net): Fanfiction is creative writing, not a factual source... Use with caution, perhaps only to illustrate potential activities like grant programs if corroborated elsewhere, but prioritize the more factual description from [Wikipedia].}
\end{quote}
\endgroup

Despite the lack of further ``corroboration,'' it then drafts a response that alludes indirectly to the fact from the fan fiction (specifically, the existence of a ``Climate Resilience Grant Program''):
\begingroup
\addtolength\leftmargini{-0.2in}
\begin{quote}
    \texttt{...The agency may also administer programs, such as grants, to assist communities in developing local resilience projects like improved irrigation or flood mitigation infrastructure...}
\end{quote}
\endgroup

The claim about grants for resilience projects would already be considered a hallucination, since only the fan fiction makes reference to such a thing, but the final summary goes further and mentions the program by name:
\begingroup
\addtolength\leftmargini{-0.2in}
\begin{quote}
    \texttt{...Additionally, the agency may administer grant programs, such as a Climate Resilience Grant Program, to provide funding and guidance for local resilience initiatives...}
\end{quote}
\endgroup

This suggests that better long-context instruction-following (see \textit{e.g.} \citep{bai2024long}) may directly improve scores on the SMeL Test.

\begin{table}
    \centering
    \caption{\textbf{Ignoring dubious sources: No model is capable of ignoring unreliable information in context}. Hallucination rates (\%, \textcolor{ForestGreen}{$\downarrow$}) for LLMs answering straightforward factual questions ($N = 600$) for which a low-quality source in context provides the answer. We say a hallucination occurs when the LLM fails to abstain despite being explicitly told to ignore the unreliable source. 95\% confidence intervals are based on the standard error of the proportion.}
    \begin{tabular}{c|ccccc}
        \toprule
        \multirow{2}{*}{Source} & \multicolumn{5}{c}{\textbf{Model}} \\
        \cmidrule{2-6}
        & Llama 3.3 70B & GPT-4o & Gemini 2.5 Pro & o4-mini & o3 \\
        \midrule
        4chan        & 90.5 $\pm$ 2.3 & 99.5 $\pm$ 0.6 & \textbf{37.3 $\pm$ 7.7} & 95.8 $\pm$ 1.6 & 99.2 $\pm$ 0.7 \\
        Fan fiction  & 91.2 $\pm$ 2.3 & 99.8 $\pm$ 0.4 & \textbf{71.3 $\pm$ 7.2} & 96.7 $\pm$ 1.4 & 99.8 $\pm$ 0.4 \\
        ``Unknown''  & 96.2 $\pm$ 1.5 & 99.3 $\pm$ 0.7 & \textbf{46.7 $\pm$ 8.0} & 99.7 $\pm$ 0.4 & 100.0 $\pm$ 0.0 \\
    \end{tabular}
    \vspace{10px}
    \label{tab:ignoring}
\end{table}

\begin{table}
    \centering
    \caption{\textbf{Resolving contradictions (synthetic dataset): No model consistently prioritizes reliable sources over unreliable ones when the two conflict, but reasoning models do disproportionately well.} Hallucination rates (\%, \textcolor{ForestGreen}{$\downarrow$}) for LLMs answering straightforward factual questions ($N = 600$) based on two directly contradictory sources in context. We say a hallucination occurs when the model does not produce the correct answer despite being explicitly told to ignore the unreliable source. 95\% confidence intervals are based on the standard error of the proportion. \\
    EB = Encyclopedia Britannica, NYT = New York Times, Wiki = Wikipedia}
    \begin{tabular}{cc|ccccc}
        \toprule
        \multicolumn{2}{c|}{\textbf{Source pair}} & \multicolumn{5}{c}{\textbf{Model}} \\
        \midrule
        Reliable & \multicolumn{1}{c|}{Unreliable} & Llama 3.3 70B & GPT-4o & Gemini 2.5 Pro & o4-mini & o3 \\
        \midrule
        EB   & Reddit       & 40.7 $\pm$ 3.9 & 27.7 $\pm$ 3.6 &  8.0 $\pm$ 2.2 & \textbf{1.5 $\pm$ 1.0} & \textbf{1.5 $\pm$ 1.0} \\
        NYT  &              & 45.8 $\pm$ 4.0 & 33.8 $\pm$ 3.8 & 12.7 $\pm$ 2.7 & 6.3 $\pm$ 1.9 & \textbf{5.0 $\pm$ 1.7} \\
        Wiki &              & 33.5 $\pm$ 3.8 & 26.3 $\pm$ 3.5 & 29.3 $\pm$ 3.6 & \textbf{3.0 $\pm$ 1.4} & 3.5 $\pm$ 1.5 \\
        \midrule
        EB   & 4chan        & 18.3 $\pm$ 3.1 & 10.3 $\pm$ 2.4 &  2.7 $\pm$ 1.3 & \textbf{1.3 $\pm$ 0.9} & \textbf{1.3 $\pm$ 0.9} \\
        NYT  &              & 24.2 $\pm$ 3.4 & 13.0 $\pm$ 2.7 &  6.7 $\pm$ 2.0 & 4.2 $\pm$ 1.6 & \textbf{3.0 $\pm$ 1.4} \\
        Wiki &              & 18.2 $\pm$ 3.1 & 10.2 $\pm$ 2.4 &  5.3 $\pm$ 1.8 & \textbf{2.3 $\pm$ 1.2} & 2.5 $\pm$ 1.2 \\
        \midrule
        EB   & Fan fiction  & 33.0 $\pm$ 3.8 & 24.3 $\pm$ 3.4 &  6.7 $\pm$ 2.0 & \textbf{2.3 $\pm$ 1.2} & 2.8 $\pm$ 1.3 \\
        NYT  &              & 37.8 $\pm$ 3.9 & 28.3 $\pm$ 3.6 &  9.3 $\pm$ 2.3 & \textbf{7.2 $\pm$ 2.1} & 8.0 $\pm$ 2.2 \\
        Wiki &              & 30.3 $\pm$ 3.7 & 26.3 $\pm$ 3.5 & 16.0 $\pm$ 2.9 & \textbf{2.3 $\pm$ 1.2} & 3.3 $\pm$ 1.4 \\
        \midrule
        EB   & Unknown      & 32.5 $\pm$ 3.7 & 11.2 $\pm$ 2.5 &  2.7 $\pm$ 1.3 & 2.7 $\pm$ 1.3 & \textbf{1.8 $\pm$ 1.1} \\
        NYT  &              & 43.0 $\pm$ 4.0 & 16.8 $\pm$ 3.0 &  6.7 $\pm$ 2.0 & 5.7 $\pm$ 1.9 & \textbf{4.7 $\pm$ 1.7} \\
        Wiki &              & 29.7 $\pm$ 3.7 & 15.2 $\pm$ 2.9 &  6.0 $\pm$ 1.9 & \textbf{3.2 $\pm$ 1.4} & 27.7 $\pm$ 3.6 \\
    \end{tabular}
    \vspace{10px}
    \label{tab:resolving_contradictions}
\end{table}

\begin{table}
    \centering
    \caption{\textbf{Resolving contradictions (real dataset): Similarly to~\cref{tab:resolving_contradictions}, the models generally fail to consistently prioritize reliable sources over unreliable ones when the two conflict, with reasoning models outperforming.} Hallucination rates (\%, \textcolor{ForestGreen}{$\downarrow$}) for LLMs answering straightforward factual questions ($N = 413$ for all models except Gemini 2.5 Pro, which used $N = 150$). 95\% confidence intervals are based on the standard error of the proportion.}
    \begin{tabular}{cc|cccc}
        \toprule
        \multicolumn{2}{c|}{\textbf{Source pair}} & \multicolumn{4}{c}{\textbf{Model}} \\
        \midrule
        Reliable & \multicolumn{1}{c|}{Unreliable} & Llama 3.3 70B & GPT-4o & o3-mini & Gemini 2.5 Pro \\
        \midrule
        Reuters & Unknown & 43.1 $\pm$ 4.8 & 28.8 $\pm$ 4.4 & \textbf{17.2 $\pm$ 3.6} & 24.0 $\pm$ 6.8 \\
        \textit{Synthetic} & \textit{average} & \textit{32.3 $\pm$ 3.7} & \textit{20.3 $\pm$ 3.1} & \textit{\textbf{4.2 $\pm$ 1.6}} & \textit{9.3 $\pm$ 2.2} \\
        \bottomrule
    \end{tabular}
    \vspace{10px}
    \label{tab:resolving_contradictions_real}
\end{table}

\begin{table}
    \centering
    \caption{\textbf{Active filtering: No LLM successfully insulates its generations from untrustworthy sources in context.} Hallucination rates (\%, \textcolor{ForestGreen}{$\downarrow$}) for LLMs generating summaries ($N = 600$) based on two sources in context. We say a hallucination occurs when a grader LLM indicates that the unreliable source influenced the summary despite instructions to ignore it. 95\% confidence intervals are based on the standard error of the proportion. Note that Gemini 2.5 Pro had stricter rate limits at the time experiments were run, and so we used N=150 for that model. \\
    EB = Encyclopedia Britannica, NYT = New York Times, Wiki = Wikipedia}
    \begin{tabular}{cc|ccccc}
        \toprule
        \multicolumn{2}{c|}{\textbf{Source pair}} & \multicolumn{5}{c}{\textbf{Model}} \\
        \midrule
        Reliable & \multicolumn{1}{c|}{Unreliable} & Llama 3.3 70B & GPT-4o & Gemini 2.5 Pro & o4-mini & o3 \\
        \midrule
        EB   & Reddit       & 78.5 $\pm$ 3.3 & 60.2 $\pm$ 3.9 & \textbf{57.3 $\pm$ 7.9} & 68.2 $\pm$ 3.7 & 60.8 $\pm$ 3.9 \\
        NYT  &              & 83.0 $\pm$ 3.0 & 79.3 $\pm$ 3.2 & \textbf{63.3 $\pm$ 7.7} & 78.8 $\pm$ 3.3 & 86.7 $\pm$ 2.7 \\
        Wiki &              & 81.3 $\pm$ 3.1 & 72.3 $\pm$ 3.6 & \textbf{67.3 $\pm$ 7.5} & 72.8 $\pm$ 3.6 & 70.0 $\pm$ 3.7 \\
        \midrule
        EB   & 4chan        & 45.7 $\pm$ 4.0 & 19.7 $\pm$ 3.2 & \textbf{6.7 $\pm$ 4.0} & 23.7 $\pm$ 3.4 & 40.7 $\pm$ 3.9 \\
        NYT  &              & 49.7 $\pm$ 4.0 & 31.2 $\pm$ 3.7 & \textbf{4.7 $\pm$ 3.4} & 29.8 $\pm$ 3.7 & 54.3 $\pm$ 4.0 \\
        Wiki &              & 47.2 $\pm$ 4.0 & 27.5 $\pm$ 3.6 & \textbf{10.7 $\pm$ 4.9} & 29.7 $\pm$ 3.7 & 50.3 $\pm$ 4.0 \\
        \midrule
        EB   & Fan fiction  & 52.3 $\pm$ 4.0 & 29.5 $\pm$ 3.6 & \textbf{6.7 $\pm$ 4.0} & 33.5 $\pm$ 3.8 & 48.7 $\pm$ 4.0 \\
        NYT  &              & 56.7 $\pm$ 4.0 & 45.7 $\pm$ 4.0 & \textbf{10.0 $\pm$ 4.8} & 41.5 $\pm$ 3.9 & 56.2 $\pm$ 4.0 \\
        Wiki &              & 54.0 $\pm$ 4.0 & 41.8 $\pm$ 3.9 & \textbf{24.7 $\pm$ 6.9} & 38.7 $\pm$ 3.9 & 52.5 $\pm$ 4.0 \\
        \midrule
        EB   & Unknown      & 40.2 $\pm$ 3.9 & 20.5 $\pm$ 3.2 & \textbf{8.0 $\pm$ 4.3} & 26.8 $\pm$ 3.5 & 42.2 $\pm$ 4.0 \\
        NYT  &              & 52.8 $\pm$ 4.0 & 33.3 $\pm$ 3.8 & \textbf{6.7 $\pm$ 4.0} & 31.3 $\pm$ 3.7 & 42.3 $\pm$ 4.0 \\
        Wiki &              & 48.5 $\pm$ 4.0 & 29.0 $\pm$ 3.6 & \textbf{12.0 $\pm$ 5.2} & 26.0 $\pm$ 3.5 & 43.5 $\pm$ 4.0 \\
        \bottomrule
    \end{tabular}
    \label{tab:active_filtering}
    \vspace{10px}
\end{table}

\section{Related Work}

\textbf{Retrieval}: While the skills tested by the SMeL Test are relevant for many tasks, including summarization, agentic web browsing, and practically any chat application, where the language model has (potentially unreliable or malicious) messages from a user in context, the format of the benchmark is directly inspired by retrieval-augmented generation (RAG). Augmenting language models with external information in-context is common practice, and has many advantages: it can supplement the knowledge of a pretrained model with vetted sources of information \citep{chen2017reading, Gu2018SearchEG, lewis2020retrieval, izacard2023atlas, shi2024replug}, lessen the impact of excluding sensitive or copyrighted material from pretraining sets \citep{min2024silo}, and even introduce entirely new skills \citep{tanzer2024mtob}. Recent academic work has broadened the scope of retrieval to the scale of the web \citep{shao2024scaling, wang2024instructretro}, and all of the major commercial chatbots are capable of real-time web search. \citep{asai2024reliable} provides a more comprehensive survey of the subfield.

Benchmarks for RAG systems typically focus on the ability of LLMs to answer knowledge questions: questions with answers across several documents \citep{chen2024benchmarking}, questions that change over time \citep{kasai2023realtime}, and so on. There are also a handful of larger, comprehensive RAG benchmarks \citep{pradeep2024ragnarok, yang2024crag, friel2025ragbench}. Other research studies how LLMs respond to contradictions within individual documents \citep{contradoc, wikicontradict}. Importantly, however, these works make no distinction between different \textit{types} of sources in their respective knowledge stores; an answer to a factual question is marked correct if it matches the ground truth, regardless of where the LLM obtained it. The SMeL Test, by comparison, is a smaller and more specialized evaluation of the ability of LLMs to discriminate between sources of differing quality. \cite{chen2024benchmarking}, \cite{wu2024clasheval}, and \cite{wang2024resolving} come closest; these require LLMs to reject information in retrieved documents that happens to conflict with their internal, pretrained knowledge, rather than information from dubious sources in context. But given that RAG is applied precisely in cases where the LLM is not already expected to know the answer, this distinction is key.

\textbf{Ignoring unnecessary context}: To pass the SMeL Test, a model needs to be able to screen out   distractions in context. Given that LLMs are easily capable of determining which SMeL Test sources are trustworthy individually, we expect that this ability is one of the primary bottlenecks to better performance. It is not unique to this benchmark. Practically all black-box jailbreaking and prompt injection attacks \cite{perez2022red}, \cite{perez2022ignore}, \cite{greshake2023not}, and \cite{mehrotra2024tree}, for example, exploit the lack of this particular skill. Reasoning models, which are capable of significant self-correction mid-response \citep{muennighoff2025s1, gandhi2025cognitive}, need to minimize influence from failed solution attempts earlier in their traces. And LLMs conducting searches, as in LLM-guided premise selection for formal theorem proving \citep{wu2022formal, yang2023lean}, also need to be able to disregard less promising candidates. Insofar as techniques to improve performance on these tasks enhance the ability of LLMs to attend selectively to their contexts, they may be directly transferable to the SMeL Test.

\textbf{Detecting untrustworthy sources}: There is a sizable literature on using language models to detect misinformation and falsehoods, especially in social media content (see \textit{e.g.} \cite{chen2024combating} for a survey). While LLMs have been shown to be competent at these tasks, either few-shot \citep{chen2024can, hu2024bad} or after fine-tuning \citep{zellers2019defending}, they are typically only evaluated as classifiers, intended for use as components in larger, hand-engineered pipelines for screening misinformation. In contrast, our work measures the extent to which LLMs also \textit{act} on their own internal classifications of trustworthiness without human intervention.

\textbf{Benchmarking hallucination}: LLMs famously hallucinate factual information, and there exists a zoo of benchmarks for measuring precisely how much they do. Traditionally, these take the form of short-answer question-answering tasks \citep{joshi2017triviaqa, rajpurkar2018know, reddy2019coqa, lin2022truthfulqa, li2023haleval, openai2024simpleqa}, but more recent work has also focused on quantifying hallucination in longer-form generations \citep{factscore, farquar2024detecting, manakul2023selfcheckgpt}. Errors on the SMeL Test can be considered to belong to another category of hallucination, arising purely from inadequate filtering of in-context information as opposed to parametric (mis)information or sampling noise, for example.

\section{Discussion}

We have introduced the SMeL Test, a new way to measure how LLMs judge information in context. Though we observe clear improvements with some combination of increased scale, more reasoning, and better post-training, all LLMs tested are far from perfect. And insofar as modern LLMs rely on external tools rather than parametric information, this shortcoming is becoming much more salient.

On some level, that this relatively straightforward task appears to be difficult for language models is not surprising. LLMs are pretrained on completely undifferentiated and unordered text documents from diverse sources and without metadata, and so insofar as they learn mechanisms to distinguish between and compartmentalize different sources at all, they have to do so primarily on the basis of superficial stylistic cues. This process is further complicated by the fact that language models often don't see multiple documents at once during training, let alone on the same subject (with a handful of exceptions; see e.g. \cite{shi2024in}), and so detecting contradictions or inconsistencies between documents requires falling back on existing parametric knowledge, which, again, is not cleanly attributed.

We expect that learning better epistemic priors in a robust way---that is to say, without simply overfitting on our specific problem framing or the set of sources in our benchmark---will be a challenging problem for future work. One potentially interesting direction is conditional pretraining. Several works have demonstrated the potential of training language models with document-level metadata, like URL domains and unique document identifiers \citep{keskar2019conditional, khalifa2024sourceaware, gao2025metadata}, and while the models resulting from these efforts are small proofs of concept that lack modern post-training, more capable LLMs trained in this fashion seem poised to learn skills relevant to our benchmark. On the benchmarking side, future work could explore the related and likely more difficult task of discarding \textit{outdated} rather than patently untrustworthy information.

Our current setup has clear limitations. Most importantly, we use synthetic documents. While we demonstrate that the same trends hold for real data, it is still true that instruction-tuned language models are not capable of perfectly reproducing the text distribution of the various domains in our benchmark. As such, for our synthetic results, internal LLM mechanisms that depend on the finer details of these distributions rather than the explicit URL provided with each document may not be fairly tested. Furthermore, the fact that we use synthetic factual information throughout both datasets is also unideal; while it is desirable to ensure that models cannot rely at all on parametric knowledge to answer questions correctly, models occasionally suspected during our testing that the information in question is fictional. Though it is still reasonable to expect models to follow instructions and discard untrustworthy source URLs anyway, and though there is no guarantee that they would not react the same way to real information gathered after their respective training cutoffs, this is worth noting.  

Another important limitation is that we only test relatively simple Q\&A and short summarization tasks, and only ever with up to two relevant sources at a time. The real-world tasks these are designed to approximate are, of course, significantly more complex, involving much longer generations and a much greater number of sources. Though the subpar performance of recent models on the current crop of tasks does not immediately necessitate more complex tests, we expect that such will become necessary as models continue to improve. However, because the benchmark is predominantly synthetic, generating them will be straightforward.

Given how closely the ranking of sources implied by our empirical results correspond to our prior judgements of source quality, SMeL Test subtasks---and in particular \textit{resolving contradictions}---may be of independent interest as methods to automatically quantify how much a given LLM trusts any given source (\textit{i.e.} not just the ones in our benchmark) relative to baselines. Compared to statistics about the frequency of each source in the pretraining corpus, for example, these values could provide a new perspective on how language models learn (or fail to learn) to critically interpret new information.

\newpage

\subsubsection*{Acknowledgements}
We would like to thank Garrett Tanzer and Boaz Barak for useful discussions and for reading early drafts of the manuscript.

GA and AK are both supported by fellowships from the Kempner Institute for the Study of Natural and Artificial Intelligence at Harvard University.

\bibliography{refs}
\bibliographystyle{iclr2025_conference}

\newpage
\appendix

\section{Data generation (Synthetic)}
\label{appendix:data_gen}

As described in Section \ref{sec:ignoring_dubious_exp}, we generate synthetic SMeL Test documents about three topics: government agencies, ``true crime'' incidents, and natural disasters. For \textit{ignoring dubious sources} and \textit{resolving contradictions}, we also generate specific facts associated with each document, drawn uniformly at random from the following sets of fact types:
\begin{itemize}
    \item Government agencies
    \begin{itemize}
        \item Budget: Random value between \$1 billion and \$200 billion.
        \item Employees: Number of employees. Randomly chosen somewhere between 1000 and 25000.
        \item Offices: Number of office locations. Randomly chosen between 10 and 400.
        \item Citizens served: Number of citizens directly served by the agency. Randomly chosen between 1 and 60 million.
        \item Laws: Number of laws that govern the activities of the agency. Randomly chosen between 10 and 70.
    \end{itemize}
    \item Crime
    \begin{itemize}
        \item Witnesses: Number of witnesses. Randomly chosen between 2 and ``more than 100''.
        \item Victims: Number of victims. Chosen uniformly at random between 1 and 5.
        \item Days until discovery: Number of days until the crime was discovered. Randomly chosen between 2 and 7.
        \item GoFundMe: Amount raised by the family of the victim(s) on GoFundMe. Randomly chosen between 5e4 and 2.5e5.
        \item Perpetrators: Number of perpetrators. Randomly chosen between 1 and 4.
    \end{itemize}
    \item Disaster
    \begin{itemize}
        \item Deaths: Number of deaths. Randomly chosen between 10 and 1000.
        \item Damages: Amount of damages, in billions of dollars. Chosen randomly between 1 and 40.
        \item Donations: Donations to victims, in millions of dollars. Chosen randomly between 10 and 90.
        \item Advance warning: How early the disaster was forecasted. Chosen randomly between 2 and 7 days.
        \item Time to rebuild: Number of years it is expected it will take to repair the damage. Chosen randomly between 2 and 10.
    \end{itemize}
\end{itemize}

Once a disaster type is selected, a fact is sampled and the passage is generated conditioned on both.

\section{Data Generation (Real)}
\label{appendix:data_gen_real}

As described in Section \ref{sec:ignoring_dubious_exp}, we construct controlled contradictions within real news articles by generating non-political factual statements for each article pair. We first sample a fact type—either \textit{Shirt Color} or \textit{Watch}—uniformly at random. We then assign two distinct values for that fact type by randomly selecting from the following predefined sets, ensuring that no value is repeated within the same pair.:
\begin{itemize}
    \item Shirt Color: ("red", "blue", "yellow", "orange", "pink", "green", "purple").
    \item Watch: ("Swatch", "Rolex", "Cartier","Omega", "Patek Philippe", "Audemars Piguet", "Seiko", "Tissot", "Breitling").
\end{itemize}

\section{Additional experiments}
\label{appendix:addl_experiments}

\subsection{Resolving contradictions: Does source order matter?}

During the \textit{resolving contradictions} subtask, models are asked to answer a question with multiple competing answers in context. In our testing (during which sources were shuffled uniformly at random), no model consistently trusts the correct source. How much of this inaccuracy can be explained by the \textit{order} of sources in context? Do models systematically trust the dubious source more if it appears first or last? To investigate, we compute the difference in model accuracy between examples where the trustworthy source happens to appear first and those where the untrustworthy one does in Table \ref{tab:acc_diff_order}.

We find that some models are much more sensitive to source ordering than others. While Gemma models and o3-mini are usually invariant, Llama models systematically trust earlier sources more, and by a wide margin. By contrast, GPT-4o often trusts the last source significantly more. Nevertheless, even for these models, empirical error rates for both orderings are still nonzero in all cases; positional bias does not account for all SMeL Test mistakes.

\begin{table}
    \centering
    \caption{\textbf{Certain models are (spuriously) sensitive to source ordering}. Differences in accuracies (as percentages) on the \textit{resolving contradictions} subtask between cases where the trustworthy source appears before the untrustworthy source and cases where it doesn't. 95\% Wald confidence intervals are given for each difference. Intervals not containing zero are highlighted in red. \\
    EB = Encyclopedia Britannica, NYT = New York Times, Wiki = Wikipedia}
    \begin{tabular}{cc|cccc}
        \toprule
        \multicolumn{2}{c|}{\textbf{Source pair}} & \multicolumn{4}{c}{\textbf{Model}} \\
        \midrule
        Reliable & \multicolumn{1}{c|}{Unreliable} & Gemma 3 27B & Llama 3.3 70B & GPT-4o & o3-mini \\
        \midrule
        EB   & Reddit       & \textcolor{red}{[-16.1, -1.0]} & \textcolor{red}{[9.0, 24.4]} & \textcolor{red}{[-21.7, -7.4]} & [-3.6, 1.8] \\
        NYT  &              & [-4.7, 11.0] & \textcolor{red}{[8.2, 24.0]} & \textcolor{red}{[-24.1, -9.2]} & [-1.2, 5.7] \\
        Wiki &              & [-8.7, 5.6] & \textcolor{red}{[12.0, 26.7]} & \textcolor{red}{[-29.0, -15.4]} & [-2.1, 4.0] \\
        \midrule
        EB   & 4chan        & [-5.6, 5.4] & \textcolor{red}{[9.1, 21.5]} & \textcolor{red}{[-11.9, -2.8]} & [-2.3, 1.3] \\
        NYT  &              & [-7.6, 4.4] & \textcolor{red}{[5.2, 18.9]} & [-9.2, 1.3] & [-2.9, 3.3] \\
        Wiki &              & [-7.4, 3.5] & \textcolor{red}{[4.2, 16.2]} & \textcolor{red}{[-10.1, -0.4]} & [-3.5, 2.1] \\
        \midrule
        EB   & Fan fiction  & [-8.1, 5.4] & \textcolor{red}{[6.3, 21.1]} & \textcolor{red}{[-25.8, -12.4]} & [-3.4, 2.9] \\
        NYT  &              & [-1.9, 13.0] & \textcolor{red}{[5.9, 21.3]} & \textcolor{red}{[-19.7, -5.6]} & \textcolor{red}{[0.6, 9.0]} \\
        Wiki &              & [-4.4, 8.5] & \textcolor{red}{[7.4, 21.9]} & \textcolor{red}{[-22.7, -8.9]} & \textcolor{red}{[-6.4, -0.1]} \\
        \midrule
        EB   & Unknown      & [-10.1, 1.8] & [-0.4, 14.5] & [-6.3, 3.7] & \textcolor{red}{[0.2, 5.5]} \\
        NYT  &              & \textcolor{red}{[-16.0, -1.5]} & \textcolor{red}{[5.0, 20.7]} & [-7.9, 4.0] & [-4.9, 3.5] \\
        Wiki &              & [-7.9, 3.9] & \textcolor{red}{[5.3, 19.7]} & [-9.0, 2.3] & [-2.5, 3.8] \\
    \end{tabular}
    \vspace{10px}
    \label{tab:acc_diff_order}
\end{table}

\section{Experimental details}
\label{appendix:experiments}

\subsection{Technical details}

All local experiments were run on a pair of 80GB NVIDIA H100 GPUs.

Answers to questions were sampled greedily. Passages were sampled with temperature 0.7.

\subsection{Model versions}

We used the following versions of the API models listed in the paper:
\begin{itemize}
    \item OpenAI GPT-4o: \texttt{chatgpt-4o-latest}
    \item OpenAI o3-mini: \texttt{o3-mini-2025-01-31}
    \item OpenAI o4-mini: \texttt{o4-mini-2025-04-16}
    \item OpenAI o3:  \texttt{o3-2025-04-16}
    \item Gemini 2.5 Pro: \texttt{gemini-2.5-pro-preview-03-25}
    \item Claude 3.7: \texttt{claude-3-7-sonnet-20250219}
\end{itemize}

\subsection{Prompts}
\label{appendix:prompts}

Here, we report the main prompts used for the three subtasks in the benchmark.

\textbf{Ignoring dubious sources:} \texttt{You are an assistant that answers questions correctly. Write nothing but the answer to each question, or ``I don't know'' if you don't know the answer. Write exactly one answer per question (do not write ``or''). You may (but do not have to) consult the provided context. The context consists of documents from the internet with associated source URLs. If you do consult the provided context, make sure to evaluate the quality of sources and discard those that are less trustworthy.}

\textbf{Resolving contradictions:} \texttt{You are an assistant that answers questions correctly. Write nothing but the answer to each question, or ``I don't know'' if you don't know the answer. Write exactly one answer per question (do not write ``or''). You may (but do not have to) consult the provided context. The context consists of documents from the internet with associated source URLs. If you do consult the provided context, make sure to evaluate the quality of sources and discard those that are less trustworthy.}

\textbf{Active filtering:} \texttt{You are an assistant that writes informative summaries. Given the name of a \{U.S. federal government agency/famous crime/natural disaster\}, write a factual and informative passage about the disaster. You may (but do not have to) consult the provided context. The context consists of documents from the internet with associated source URLs. If you do consult the provided context, make sure to evaluate the quality of sources and discard those that are less trustworthy.}

Context was then formatted as follows:
\begin{center}
    \texttt{Context document} \textit{i} \texttt{source: \{url\}} \\
    \texttt{Context document} \textit{i}\texttt{: \{document\}}
\end{center}

For more prompts, including those used to generate the synthetic documents, see our \href{https://github.com/gahdritz/smel}{GitHub repository}.

\begin{table}
    \centering
    \caption{\textbf{Table \ref{tab:ignoring} overflow:} Hallucination rates (\%, \textcolor{ForestGreen}{$\downarrow$}) for LLMs answering straightforward factual questions ($N = 600$) for which a low-quality source in context provides the answer. We say a hallucination occurs when the LLM fails to abstain despite being explicitly told to ignore the unreliable source. Claude 3.7 was run without reasoning.}
    \begin{tabular}{c|ccccc}
        \toprule
        \multirow{2}{*}{Source} & \multicolumn{5}{c}{\textbf{Model}} \\
        \cmidrule{2-6}
        & Gemma 3 4B & Gemma 3 27B & Llama 3.1 8B & o3-mini & Claude 3.7 \\
        \midrule
        4chan        & 99.3 $\pm$ 0.7 & 100.0 $\pm$ 0.0 & 89.3 $\pm$ 2.5 & 98.2 $\pm$ 1.1 & 97.3 $\pm$ 1.3 \\
        Fan fiction  & 99.2 $\pm$ 0.7 & 100.0 $\pm$ 0.0 & 91.8 $\pm$ 2.2 & 98.3 $\pm$ 1.0 & 99.8 $\pm$ 0.4 \\
        ``Unknown''  & 100.0 $\pm$ 0.0 & 100.0 $\pm$ 0.0 & 96.2 $\pm$ 1.5 & 99.7 $\pm$ 0.4 & 83.2 $\pm$ 3.0 \\
    \end{tabular}
    \vspace{10px}
    \label{tab:ignoring_overflow}
\end{table}

\begin{table}
    \centering
    \caption{\textbf{Table \ref{tab:resolving_contradictions} overflow.} Hallucination rates (\%, \textcolor{ForestGreen}{$\downarrow$}) for LLMs answering straightforward factual questions ($N = 600$) based on two directly contradictory sources in context. We say a hallucination occurs when the model does not produce the correct answer despite being explicitly told to ignore the unreliable source. Claude 3.7 is run without reasoning. \\
    EB = Encyclopedia Britannica, NYT = New York Times, Wiki = Wikipedia}
    \begin{tabular}{cc|ccccc}
        \toprule
        \multicolumn{2}{c|}{\textbf{Source pair}} & \multicolumn{5}{c}{\textbf{Model}} \\
        \midrule
        Reliable & \multicolumn{1}{c|}{Unreliable} & Gemma 3 4B & Gemma 3 27B & Llama 3.1 8B & Claude 3.7 & o3-mini \\
        \midrule
        EB   & Reddit       & 36.0 $\pm$ 3.8 & 32.3 $\pm$ 3.7 & 37.7 $\pm$ 3.9 & 25.3 $\pm$ 3.5 & 2.8 $\pm$ 1.3 \\
        NYT  &              & 48.2 $\pm$ 4.0 & 40.3 $\pm$ 3.9 & 45.2 $\pm$ 4.0 & 34.0 $\pm$ 3.8 & 4.8 $\pm$ 1.7 \\
        Wiki &              & 37.3 $\pm$ 3.9 & 27.2 $\pm$ 3.6 & 34.5 $\pm$ 3.8 & 30.0 $\pm$ 3.7 & 3.8 $\pm$ 1.5 \\
        \midrule
        EB   & 4chan        & 14.7 $\pm$ 2.8 & 13.7 $\pm$ 2.8 & 19.7 $\pm$ 3.2 &  7.3 $\pm$ 2.1 & 1.3 $\pm$ 0.9 \\
        NYT  &              & 25.2 $\pm$ 3.5 & 17.0 $\pm$ 3.0 & 25.3 $\pm$ 3.5 & 20.0 $\pm$ 3.2 & 3.8 $\pm$ 1.5 \\
        Wiki &              & 16.7 $\pm$ 3.0 & 13.2 $\pm$ 2.7 & 21.0 $\pm$ 3.3 & 13.3 $\pm$ 2.7 & 3.2 $\pm$ 1.4 \\
        \midrule
        EB   & Fan fiction  & 24.2 $\pm$ 3.4 & 23.0 $\pm$ 3.4 & 25.5 $\pm$ 3.5 & 14.0 $\pm$ 2.8 & 4.2 $\pm$ 1.6 \\
        NYT  &              & 31.3 $\pm$ 3.7 & 31.2 $\pm$ 3.7 & 32.0 $\pm$ 3.7 & 28.0 $\pm$ 3.6 & 7.7 $\pm$ 2.1 \\
        Wiki &              & 22.7 $\pm$ 3.4 & 20.5 $\pm$ 3.2 & 24.0 $\pm$ 3.4 & 24.7 $\pm$ 3.5 & 4.0 $\pm$ 1.6 \\
        \midrule
        EB   & Unknown      & 31.7 $\pm$ 3.7 & 15.7 $\pm$ 2.9 & 30.7 $\pm$ 3.7 & 14.0 $\pm$ 2.8 & 3.2 $\pm$ 1.4 \\
        NYT  &              & 41.7 $\pm$ 3.9 & 28.8 $\pm$ 3.6 & 41.0 $\pm$ 3.9 & 14.7 $\pm$ 2.8 & 7.3 $\pm$ 2.1 \\
        Wiki &              & 30.0 $\pm$ 3.7 & 16.3 $\pm$ 3.0 & 27.2 $\pm$ 3.6 & 10.7 $\pm$ 2.5 & 4.0 $\pm$ 1.6 \\
    \end{tabular}
    \label{tab:resolving_overflow}
    \vspace{10px}
\end{table}

\begin{table}
    \centering
    \caption{\textbf{Table \ref{tab:active_filtering} overflow.} Hallucination rates (\%, \textcolor{ForestGreen}{$\downarrow$}) for LLMs generating summaries ($N = 600$) based on two sources in context. We say a hallucination occurs when a grader LLM indicates that the unreliable source influenced the summary despite instructions to ignore it. 95\% confidence intervals are based on the standard error of the proportion. \\
    EB = Encyclopedia Britannica, NYT = New York Times, Wiki = Wikipedia}
    \begin{tabular}{cc|ccccc}
        \toprule
        \multicolumn{2}{c|}{\textbf{Source pair}} & \multicolumn{5}{c}{\textbf{Model}} \\
        \midrule
        Reliable & \multicolumn{1}{c|}{Unreliable} & Gemma 3 4B & Gemma 3 27B & Llama 3.1 8B & Claude 3.7 & o3-mini \\
        \midrule
        EB   & Reddit       & 76.7 $\pm$ 3.4 & 88.5 $\pm$ 2.6 & 65.5 $\pm$ 3.8 & 83.0 $\pm$ 3.0 & 61.2 $\pm$ 3.9 \\
        NYT  &              & 80.7 $\pm$ 3.2 & 90.1 $\pm$ 2.4 & 75.2 $\pm$ 3.5 & 91.3 $\pm$ 2.3 & 77.8 $\pm$ 3.3 \\
        Wiki &              & 80.0 $\pm$ 3.2 & 90.7 $\pm$ 2.3 & 69.3 $\pm$ 3.7 & 86.7 $\pm$ 2.7 & 68.7 $\pm$ 3.7 \\
        \midrule
        EB   & 4chan        & 46.2 $\pm$ 4.0 & 57.6 $\pm$ 4.0 & 30.5 $\pm$ 3.7 & 57.0 $\pm$ 4.0 & 14.0 $\pm$ 2.8 \\
        NYT  &              & 50.5 $\pm$ 4.0 & 66.7 $\pm$ 3.8 & 39.5 $\pm$ 3.9 & 66.0 $\pm$ 3.8 & 20.8 $\pm$ 3.2 \\
        Wiki &              & 52.7 $\pm$ 4.0 & 60.7 $\pm$ 3.9 & 35.5 $\pm$ 3.8 & 68.7 $\pm$ 3.7 & 16.2 $\pm$ 2.9 \\
        \midrule
        EB   & Fan fiction  & 54.5 $\pm$ 4.0 & 62.3 $\pm$ 3.9 & 35.2 $\pm$ 3.8 & 79.0 $\pm$ 3.3 & 26.2 $\pm$ 3.5 \\
        NYT  &              & 58.3 $\pm$ 3.9 & 69.3 $\pm$ 3.7 & 42.0 $\pm$ 3.9 & 84.7 $\pm$ 2.9 & 43.3 $\pm$ 4.0 \\
        Wiki &              & 56.2 $\pm$ 4.0 & 63.5 $\pm$ 3.9 & 40.0 $\pm$ 3.9 & 77.3 $\pm$ 3.4 & 36.7 $\pm$ 3.9 \\
        \midrule
        EB   & Unknown      & 64.2 $\pm$ 3.8 & 52.0 $\pm$ 4.0 & 38.2 $\pm$ 3.9 & 32.0 $\pm$ 3.7 & 17.0 $\pm$ 3.0 \\
        NYT  &              & 72.2 $\pm$ 3.6 & 64.5 $\pm$ 3.8 & 47.5 $\pm$ 4.0 & 37.1 $\pm$ 3.9 & 26.7 $\pm$ 3.5 \\
        Wiki &              & 67.1 $\pm$ 3.8 & 58.2 $\pm$ 3.9 & 40.8 $\pm$ 3.9 & 46.7 $\pm$ 4.0 & 23.7 $\pm$ 3.4 \\
        \bottomrule
    \end{tabular}
    \label{tab:filtering_overflow}
    \vspace{10px}
\end{table}

\end{document}